\documentclass[10pt]{article}
    \usepackage[preprint]{rlj}
    \usepackage{amssymb}
    \usepackage{graphicx}
    \usepackage{subcaption}
    \usepackage{url}
    \usepackage{booktabs}

    \title{Dynamic Object Masks as Goal Representations for Visual Goal-Conditioned Reinforcement Learning}
    \author{Fahim Shahriar\textsuperscript{1,4}, Cheryl Wang\textsuperscript{2}, Alireza Azimi\textsuperscript{1,4}, Gautham Vasan\textsuperscript{1,4}, Hany Hamed\textsuperscript{1,4,8}, Abhishek Naik\textsuperscript{5}, A.\ Rupam Mahmood\textsuperscript{1,4,7}, Colin Bellinger\textsuperscript{3,6,*}}

    \emails{\{fshahri1, sazimi, vasan, hhamed1, armahmood\}@ualberta.ca, \ huiyi.wang@mail.mcgill.ca, \ abhishek.naik@nrc-cnrc.gc.ca, \  colin.bellinger@uottawa.ca}

    \affiliations{
    $^{1}$\textbf{University of Alberta}\\
    $^{2}$\textbf{McGill University}\\
    $^{3}$\textbf{University of Ottawa}\\
    $^{4}$\textbf{Alberta Machine Intelligence Institute (Amii)}\\
    $^{5}$\textbf{National Research Council Canada}\\
    $^{6}$\textbf{Vector Institute}\\
    $^{7}$\textbf{CIFAR Canada AI Chair}\\
    $^{8}$\textbf{Generative Bionics}\\
    $^{*}$Work done while at the National Research Council Canada
    }

\begin{document}
    \maketitle

    \begin{abstract}
    Goal-conditioned reinforcement learning (GCRL) offers a unified way to pursue diverse tasks, yet most existing methods rely on state- or position-based goal representations that are unavailable in real-world robotics. Robots operating in warehouses, agriculture, or laboratory environments rarely have access to privileged goal states, object positions, or future observations, limiting the practicality of current GCRL approaches.
    We propose a dynamic mask-based goal representation that provides simple, object-agnostic visual cues for vision-based navigation and manipulation. At each timestep, an image-based goal detector produces a goal mask using standard image processing, task-specific object recognizers, or pretrained detectors such as Detic or Grounding DINO. These masks specify the spatial target without requiring privileged goal states, enabling broad applicability and strong generalization to unseen objects.
    Our method improves stability and sample efficiency in GCRL, achieving a 99\% success rate in reaching both training and novel objects with Franka and UR10e robotic arms and faster learning in simulated navigation. We further demonstrate learning from scratch and sim-to-real transfer on both robotic arms, highlighting the effectiveness of our approach for real-world RL tasks. Our code is available at \url{https://github.com/fahimfss/GCRL}.

    \end{abstract}
    
    \section{Introduction} \label{sec:intro}
    
    Many real-world robotic tasks, such as sorting objects in e-commerce warehouses, fruit picking, or visual navigation to a target site, involve solving a multi-goal problem. These tasks require the agent to act in a particular way among numerous options to achieve various desired outcomes. Goal Conditioned Reinforcement Learning (GCRL) enables multi-goal learning by allowing the agent to acquire the necessary skills to solve a range of objectives, using a unified policy.

    Goal representation plays a key role in GCRL, affecting essential components of robot learning such as convergence speed, learned behavior, and generalization capabilities \citep{schaul15, Liu2022}. Much of the GCRL literature relies on representations that assume access to privileged state information or desired future observations, such as vectorized position or orientation and target state images \citep{Andrychowicz2017, Plappert2018, Nair2018, Liu2022}. In many real-world robotic tasks, however, such information is not readily available. Researchers adapting GCRL to real-world manipulation have used representations such as 3D target position \citep{Plappert2018}, VAE-encoded goal images \citep{Nair2018, cong2022reinforcement}, and text-based embeddings \citep{homerobot}. While target position provides information-rich goal conditions that enable efficient learning, obtaining it requires specialized depth cameras and processing techniques such as feature matching and outlier rejection \citep{dong2022towards}. The accuracy of these estimates is significantly impacted by the distance, reflectivity, and transparency of the target \citep{thalhammer2024challenges}, as well as unpredictable lighting conditions \citep{al2025enhancing}. Image- and text-based representations are easier to apply, but we find they limit the agent's ability to generalize to new targets (Section \ref{sec:results}).
    
    We propose a mask-based goal representation that is well-suited for multi-goal and visual robotic reaching, manipulation, and navigation tasks. Our method adds a binary channel to the agent's visual input specifying the target. The mask is updated at each time step to provide the agent with object-agnostic visual cues and progress toward the target. Our experiments demonstrate that this representation matches or outperforms existing methods on visual reaching tasks with both seen and unseen targets, showing good generalization. Complementary to prior work on exploration in GCRL, we focus on goal representation and reward design for real-world robotic tasks, aiming to reduce reliance on privileged 3D information and enable faster, more reliable learning. 
    
    Beyond goal representation, reward design is an equally important challenge in GCRL. Sparse rewards are particularly difficult, since the agent may need to execute a long sequence of correct actions before receiving positive feedback \citep{vasan2024revisiting}. This leads to poor sample efficiency and the failure to learn complex tasks. Dense reward, particularly distance-to-goal-based reward, is commonly used to mitigate this problem \citep{Trott}. However, it often introduces new local optima that strongly depend on the environment and task definition, ultimately preventing agents from learning the optimal behavior \citep{Trott}.

    In this work, we show that mask size can be used as an effective dense reward signal, since the target mask changes size and location based on the position of the target throughout the episode (Figure \ref{fig:pickup}). Using mask-based reward removes the dependency on computing the target's 3D world position, significantly simplifying real-world implementation. Mask-based goal conditioning and reward signals form a novel combination that, to our knowledge, has not been explored in prior work. In contrast to using masks only as a fixed goal representation, our mask is updated over time to track the target and provide a dense progress signal via its size. Our approach focuses on egocentric tasks, where the robot perceives the environment from its onboard camera as it physically moves toward the target. The mask-based goal representation works with any camera setup, but the mask-size reward assumes an egocentric view, where the mask grows as the robot approaches the target. For other camera configurations, the mask can still be incorporated into the reward signal in a task-specific way.

    We demonstrate that pretrained object detection models such as Detic \citep{detic_zhou2022detecting} and Grounding DINO \citep{GroundingDINO} can be used for mask generation in real-world tasks. We show the feasibility of using these pretrained models on two real-world applications:  a) UR10e arm-based sim-to-real transfer, and b) learning from scratch using the Franka Panda arm. For the sim-to-real setup, we additionally use a novel image interpolation technique in training to mitigate visual domain shift during real-world policy transfer.
    
    Our main contributions include: (1) the introduction of object masking as a goal representation strategy in vision-based GCRL; (2) an effective mask-based reward signal that minimizes the need for privileged information; (3) performing visual pick-up utilizing region-of-interest-based masks; (4) demonstrating the effectiveness of the proposed method in sim-to-real and real-world learning from scratch tasks.

    \section{Related Work}
    \label{sec: Related Work}
    
    \subsubsection*{\textbf{Goal Representations}} Vectorized position or orientation of an object or the robot is a common goal representation system used in the GCRL literature \citep{Plappert2018, Florensa2017, Tang2020}. Other vectorized goal representations include one-hot encoding \citep{one_hot_enc_1, one_hot_enc_2}, and object text description to CLIP embeddings \citep{homerobot}. For visual tasks, a common approach is to use variational autoencoders (VAE) to encode visual inputs such as RGB images into a latent space \citep{Nair2018, cong2022reinforcement}. However, traditional image-based goal conditioning tasks require a predefined goal state, which is unfeasible for tasks where such conditions are unknown and require exploration by the agent. In contrast, masks are simpler to generate and provide direct state-dependent feedback to the agent. 
    
    Mask-based Goal Representation: 
    Object masking has been applied by researchers to represent target objects and goal positions under stationary camera configurations \citep{wang2021roll, cong2022reinforcement}. These approaches train VAEs to encode images into latent vectors prior to GCRL training and use a complex latent state-based reward. On the other hand, our approach learns directly from images, uses masks to track progression towards the target, and uses mask-size-based rewards, which greatly simplifies the learning system.
    
    \citet{homerobot} used segmentation masks and depth data for target object finding and reaching tasks. \citet{stone2023} used masks with single activated pixels to denote target objects, allowing good generalization to objects of any shape and unseen objects. \citet{mendonca2024} used object masks to obtain the target object's point-cloud information to perform mobile manipulation tasks. Our proposed system uses dynamic masks to achieve faster convergence and utilizes mask-based dense reward signals, which none of the approaches mentioned above employ.

    \subsubsection*{\textbf{Sparse Reward Challenge for GCRL}} Sparse reward-based GCRL tasks often face an efficiency issue, as it takes a very long time for the reward signal to propagate from the goal states to the starting states \citep{Andrychowicz2017, Liu2022}. One approach to tackle this challenge is the use of distance-based dense rewards in either Euclidean space or latent space \citep{Nair2018, Plappert2018, cong2022reinforcement}. Nevertheless, this simple reward shaping is vulnerable to local optima that stem from a poor exploration of the environment \citep{vasan2024revisiting}. Alternatively, hindsight experience replay (HER) addresses the sparse reward issue by re-assigning failed states as alternative goals and providing denser reward signals to speed up learning \citep{Andrychowicz2017}. However, the HER-based approach still suffers from sample efficiency in complex environments \citep{luo2022relay}. For vision-based tasks, imagined sub-goals between the current and final goal state can simplify learning for long-horizon tasks \citep{Nair2018, Chane-Sane2021}, but these predictions sometimes result in infeasible robotic states or unrealistic images.

    \subsubsection*{\textbf{Object Detection Models in Robotic Tasks}} 
    For vision-based robotic tasks, a common practice is to use object detection models to locate objects of interest. YOLO models are a common choice for researchers, due to their fast inference times and good accuracy \citep{Spin24, zhang2023cherry}. However, YOLO models support a fixed number of predefined object detections. As a result, users are required to retrain or fine-tune YOLO models to support novel objects \citep{yolo_ft_1, yolo_ft_2}. Recently, open vocabulary-based object detection models have gained popularity in the robotics community due to zero-shot detection capabilities. Open vocabulary object detectors, such as Detic and Grounding DINO, have been used in various robotic manipulation tasks \citep{mendonca2024, stone2023, homerobot}. In this work, we utilize pretrained open-vocabulary detection models to generate masks for goal conditioning and reward computation and to assess their feasibility in real-world tasks.

    \section{Background}
      
    In standard RL, the objective is to train a policy to maximize the expected return. GCRL policies have a similar objective, but the expected return in each episode is conditioned on a predetermined goal $g \in \mathcal{G}$. GCRL can be formally represented by the goal-augmented Markov decision process (GA-MDP). At each time-step $t$, the agent samples an action $A_t \in \mathcal{A}$ based on the current state $S_t \in \mathcal{S}$, and the current goal $g$, using a policy $\pi_\phi$, $A_t \sim \pi_\phi(\cdot \mid S_t, g)$. The agent receives a reward $R_{t+1}$ and the next state $S_{t+1}$ at the next time-step $t+1$ based on a transition function \mbox{$P$, $S_{t+1}, R_{t+1} \sim P(\cdot,\cdot\mid S_t, A_t, g)$}. The episode ends when the agent reaches the goal, exceeds the allowed time limit, or leaves the allowed state space. 
    
    In this work, we evaluate the mask-based goal representation using on and off-policy methods. A brief overview of each method is provided below. Readers are directed to the original papers \citep{haarnoja2018, ppo}, for more details. 
    
    Soft Actor-Critic (SAC)  is widely used for vision-based robotic tasks due to its higher sample efficiency. In GCRL, SAC learns the parameters $\phi$ of a policy \mbox{$\pi_\phi$} to maximize a trade-off between the policy's entropy and expected return. SAC uses a soft-Q function (critic) $Q_\theta(S_t, A_t, g)$, parameterized by $\theta$, to estimate expected returns and the entropy bonus based on the current state, action, and goal. 
    
    Proximal Policy Optimization (PPO) is another frequently used RL algorithm for robotics tasks, particularly for sim-to-real applications. PPO updates policy parameters $\phi$ by maximizing the PPO-Clip objective and constraining changes to the new policy, improving the learning stability. PPO uses an advantage estimate $\hat{A_t}$ that denotes how good the selected action $a_t$ is in the current state $S_t$, and learns to increase the probabilities of selecting good actions. The critic function $V_t(S_t, g)$ predicts the expected return for a given state and goal, and it is used to generate accurate advantage estimates. 
    
    For vision-based GCRL tasks, the state $S_t$ generally consists of an image $I_t$ and a vector $v_t$, \mbox{$S_t = \langle I_t, v_t \rangle$}. An image encoder  $f_\psi : \mathbb{R}^{H\times W \times C} \to \mathbb{R}^d$ is used to process the image into a lower dimensional latent vector; here $\psi$ represents the learnable parameters, $d$ is the latent dimension, and $H$, $W$, and $C$ denote the height, width, and the number of channels of the input image respectively. Both the policy $\pi_\phi$ and the critic $Q_\theta$ use $f_\psi$ to process the image, but the parameter $\psi$ is only updated during the critic update.

    \vspace{4pt}
    
    \textbf{Reward Systems:} In GCRL, sparse rewards are only provided to the agent upon task completion:
    {
    \setlength{\belowdisplayskip}{2pt}
    \begin{equation}
    R^{sparse}(s_t, a_t, g) =R^{\text{term}}\quad \{ \text{goal reached} \} 
    \end{equation}
    }
    
Distance-to-goal-based dense reward systems are often used as an alternative to the sparse reward system: 
{
\small
\setlength{\abovedisplayskip}{3pt}
\setlength{\belowdisplayskip}{3pt}
\begin{equation} \label{eq:reward_dist}
R^{dist}(s_t, a_t, g) =
\begin{cases}
-d(S_{t}, g) & \text{if } d(S_{t}, g) > \epsilon, \\
-d(S_{t}, g) + R^{\text{term}} & \text{if } d(S_{t}, g) \le \epsilon
\end{cases}
\end{equation}
}
Here, $d(S_{t}, g)$ denotes the distance to the goal for the state $S_{t}$, $\epsilon$ is the minimum distance to the goal, and $R^{\text{term}}$ denotes a terminal reward. In our simulated experiments, the distance is computed between the midpoint of the gripper fingers and the reference frame of the target object.

    \begin{figure*}[t]
      \centering
      \begin{subfigure}[t]{0.22\linewidth}
        \includegraphics[width=\linewidth]{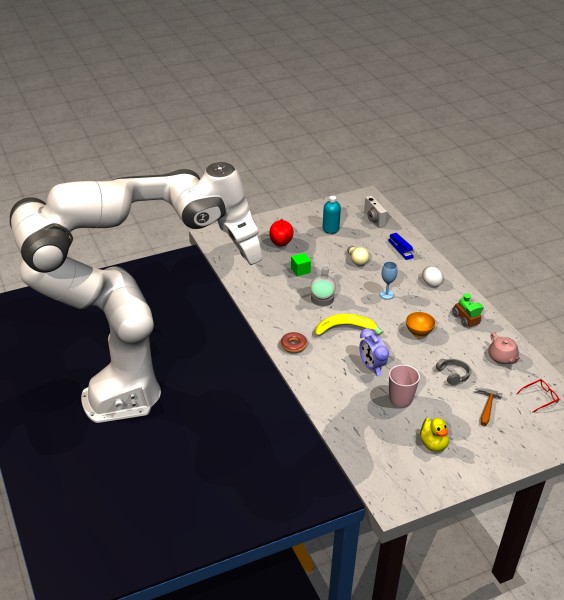}
        \caption{\footnotesize Franka-MuJoCo environment. Example object names: chocolate donut, banana, and red apple.}
        \label{fig:arm1}
      \end{subfigure}\hfill
      \begin{subfigure}[t]{0.21\linewidth}
        \includegraphics[width=\linewidth]{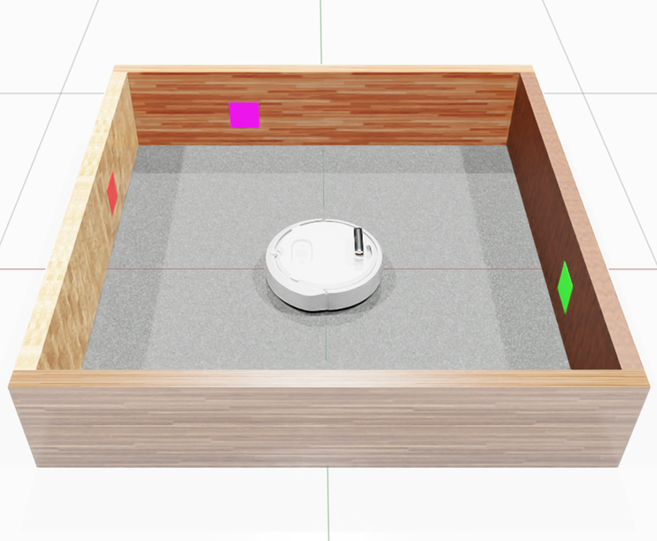}
        \caption{\footnotesize Create3-Isaac Sim environment.}
        \label{fig:create3_env}
      \end{subfigure}\hfill
      \begin{subfigure}[t]{0.18\linewidth}
        \includegraphics[width=\linewidth]{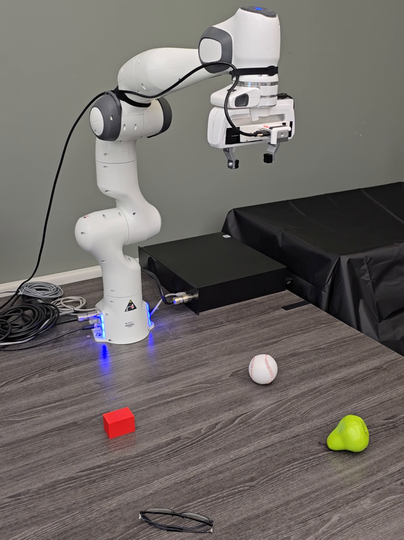}
        \caption{\footnotesize Learning from scratch setup with the Franka arm.}
        \label{fig:franka_rw}
      \end{subfigure}\hfill
      \begin{subfigure}[t]{0.18\linewidth}
        \includegraphics[width=\linewidth]{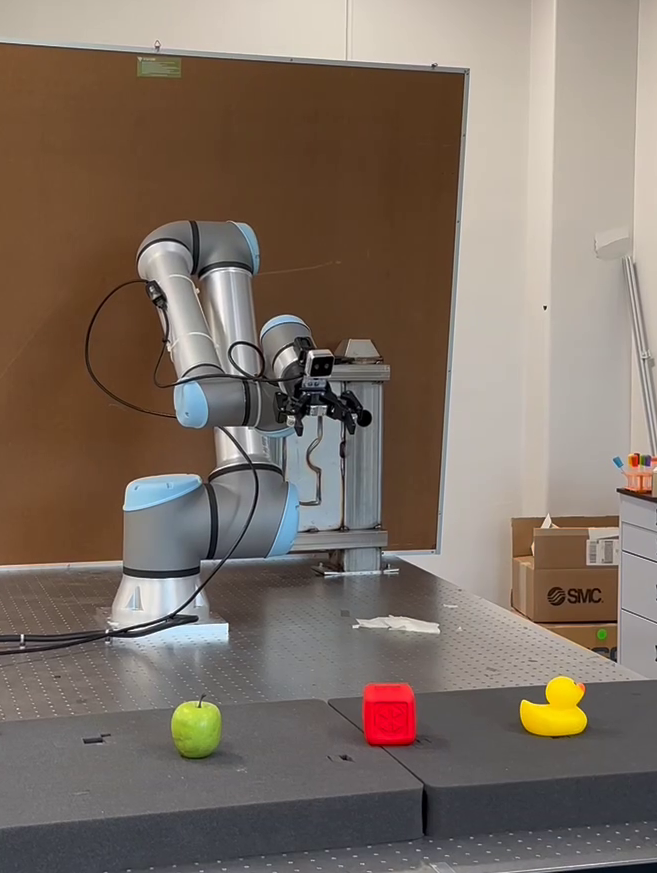}
        \caption{\footnotesize Sim-to-real setup with the UR10e arm.}
        \label{fig:arm2}
      \end{subfigure}\hfill
      \setlength{\abovecaptionskip}{-1pt} 
      \setlength{\belowcaptionskip}{-14pt}
      \caption{Simulated and real-world environments.}
      \label{fig:example1}
    \end{figure*}

    \section{Mask-based Visual GCRL}\label{sec:mask_gcrl}
    Our proposed mask-based GCRL system includes a novel mask-based goal representation and reward formulation, along with targeted algorithmic enhancements for stable visual learning.
    
    \subsection{Mask-based Goal Representation} 
    Masks are single-channel, binary images with the goal represented by a region of activated pixels. At each time step $t$, a mask $m_t$ is created using the image from a camera sensor $I_t^{cam}$. The ground truth mask can be generated using privileged positional information of the target in simulation. Alternatively, object color-based or shape-based segmentation or pretrained object detectors can be used to generate masks. The mask is then appended to $I_t^{cam}$ on the channel axis, to form $I_t' = [I_t^{cam}\ m_t]_C$. Three of the most recent such images are stacked on the channel axis to form $I_t$, where $I_t = [I_{t}' \ I_{t-1}'\ I_{t-2}']_C$, and have the shape $ 90 \times 160 \times 12 $, representing height, width, and channel. 
    
    Mask-based goal representation offers several advantages (demonstrated in Section \ref{sec:results}): (1)  the mask is calculated at each time step, providing richer signals and enabling faster training. (2) It is object-agnostic and facilitates better generalization to novel objects. (3) Masks can be generated in multiple ways, and we demonstrate three such approaches: (a) using privileged object position, (b) using object detection models, and (c) using color filtering.
    
    \subsection{Mask-based Reward Formulation}
    The mask-based reward system serves as an alternative to the distance-based reward. The mask-based reward system replaces the distance to the goal term, $-d(S_{t}, g)$, in Equation \eqref{eq:reward_dist} with $R^{mask}$:
    {
    \setlength{\abovedisplayskip}{2pt}
    \setlength{\belowdisplayskip}{4pt}
    \begin{equation} \label{eq:reward_temp}
    R' = \frac{1}{H \times W} \sum_{i=1}^{H} \sum_{j=1}^{W} \mathbf{1}\{ m_t^{ij} > 0 \}
    \end{equation}
    }
    {
    \setlength{\abovedisplayskip}{-8pt}
    \setlength{\belowdisplayskip}{10pt}
    \begin{equation} \label{eq:reward_mask}
    R^{mask} = \frac{2}{1 + e^{-10\,R'}} - 1, \quad R^{mask} \in [0,1]
    \end{equation}
    }
    Here, $H$ and $W$ are the height and width of the mask $m_t$, respectively. Equation \eqref{eq:reward_temp} computes a normalized reward, $R'$, based on the activated pixels in $m_t$. Equation \eqref{eq:reward_mask} applies a scaled sigmoid function to $R'$, which increases its value for small inputs, specifically when \mbox{$R' < 0.5$}. This scaling provides stronger feedback to the agent, particularly when the goal is further away and the changes in the mask sizes are small. Although the reward can be used without this sigmoid shaping (i.e., directly from $R'$), applying it amplifies small gains for $R' < 0.5$ and improves training speed.
    
    \subsection{Visual Learning Architectures} 
    
    {\textbf{SAC}}: Our SAC implementation utilizes asynchronous network updates and replay buffer sampling to enable efficient learning in simulation and the real world \citep{yuan2022}. To stabilize learning from pixels and to achieve better learning performance, we incorporated random shift \citep{yarats2021}, used an ensemble of five Q-Networks \citep{chen2021}, and weight clipping \citep{elsayed2024} with $\kappa = 2$ and $s_l = \sqrt{2}$ in our SAC implementation. All experiments in our work, except sim-to-real on UR10e (see Section \ref{sec:Sim2Real}), utilize SAC-based GCRL agents.  
    
    {\textbf{PPO}}: 
    For the sim-to-real application, we used the Stable Baselines 3 implementation of PPO  \citep{stable-baselines3}. We chose PPO because it is widely used in sim-to-real tasks \citep{zhao2020sim}. We used a novel image interpolation technique for training the PPO agent, which was essential for addressing visual domain shift during real-world policy transfer. More details are in Section \ref{sec:Sim2Real}.

    \section{Environments}
    We evaluate our method across three simulated environments covering both manipulation and navigation tasks. Applications to physical robots, including sim-to-real transfer and learning from scratch, are discussed in Section \ref{sec:real_world_app}.
    
    \subsection*{\textbf{Franka-MuJoCo Environment}}
    We created a tabletop setup with a Franka Panda robotic arm collected from MuJoCo Menagerie \citep{menagerie2022github}, and 20 everyday objects using \citet{RoboHive2020}, a MuJoCo-based robot learning framework. Figure \ref{fig:arm1} shows the simulated environment with all the objects placed on the tabletop. We attached a camera near the end-effector of the arm to facilitate vision-based reach, grasp, and lift. During training, three to five objects are arbitrarily selected from a pool of fifteen objects in each episode. The remaining five objects are held out as test objects. The selected objects are placed randomly on the table in front of the arm, and one object is randomly selected to be the target. 
    
    The goal of the GCRL agent for this task is to control individual joints of the Franka arm using low-level velocity commands to reach the target object within 150 time steps. The common state space across experiments contains a stack of the three most recent camera images of the shape $ 90 \times 160 \times 9 $, the joint positions, and the last action. 
    
    \subsection*{\textbf{UR10e-MuJoCo Environment}}
    For this environment, we replaced the Franka arm with the UR10e arm (MuJoCo Menagerie). All other aspects of the environment remained the same. The joint lengths, angular velocities (range and speed), and orientation of the joints for the UR10e arm are substantially different compared to the Franka arm. Furthermore, the UR10e arm is more sensitive to actions compared to the Franka arm, making the learning task slightly more difficult. 
    
    \subsection*{\textbf{Create 3-Isaac Sim Environment}}
    We created a square arena in the Isaac Sim simulator \citep{nvidia_isaac_sim_2025} and used an iRobot Create 3 mobile robot to test the visual navigation performance of the five goal conditioning methods. In each episode, we placed randomly chosen colored stickers (magenta, green, red, or blue) on each wall of the arena, and assigned a random pose to the Create 3 (Figure \ref{fig:create3_env}). The GCRL agent needs to control the linear and angular velocities of the Create 3 to reach the selected target sticker in 200 time steps. We used three stacked images and a vectorized history,  $v_t$, of the last 15 actions made by the agent. This allows the agent to have access to detailed information about the past motions of the robot. For this environment, we used color-filtering techniques to generate the masks.
    
    In addition to these environments, in Section \ref{sec:real_world_app}, we demonstrate sim-to-real transfer and learning from scratch on the UR10e and Franka Panda robots. 

    \section{Experimental Setup}\label{sec:exp_setup}

    We aim to answer the following research questions through our experiments:
\begin{enumerate}
    \item \textit{Does the mask-based system enable (a) faster training, and (b) better generalization to unseen objects?}
    \item \textit{Is mask-based reward a viable alternative to distance-based dense reward?}
    \item \textit{Is it possible to use mask-based GCRL and reward systems for performing complex tasks, such as picking up?}
\end{enumerate}

    To address these research questions more specifically, we evaluate our approach against existing goal representations.
    The \textbf{target state} goal representation adds a pre-existing close-up RGB image of the target with size $90 \times 160 \times 12 $ to the stacked image states,  $I_t$. \textbf{One-hot} encoding,  \textbf{3D position}, and \textbf{CLIP embedding} are vector-based goal representations and appended to the vector part of the state space, $v_t$. One-hot encoding uses a 20D zero vector with a single activated index that identifies the target. CLIP embedding uses a 512D vector generated using a pretrained CLIP encoder based on a text description of the target object \citep{homerobot}. 3D position uses the privileged world position of the target collected from the simulator. 
    
    We design a sequence of experiments that address each of the research questions above.
    \subsubsection*{\textbf{Experiment 1 (a)}} 
    To demonstrate the learning efficiency of the mask-based system, we trained mask-based GCRL agents in \textit{Franka-MuJoCo}, \textit{UR10e-MuJoCo}, and \textit{Create3-Isaac Sim} environments and compared the learning performance against one-hot encoding, 3D position, CLIP embeddings, and target state-based goal representation systems. Note that the mask-based rewards are not used in these experiments. We used the distance-based dense reward defined in Equation \eqref{eq:reward_dist}, with $\epsilon$ set to 10 cm, and $R^{term} = 5$. Each experiment used eleven seeds, and we trained each agent for 300,000 steps. We used a replay buffer of size 300,000, and the initial 5,000 steps were used to pre-fill it. 
    
    \subsubsection*{\textbf{Experiment 1 (b)}} We tested the reaching capabilities of the agents trained in \textit{Experiment 1 (a)} for both seen and unseen objects for the \textit{Franka-MuJoCo} and \textit{UR10e-MuJoCo} environments. We ensured similarity across the experiments by selecting the five test objects deterministically based on the seed. To collect the reaching accuracy, for each seed, we carried out twenty-five trials for each object, resulting in a total of 5,500 trials for each goal representation system. A trial is considered successful when the end-effector reaches within $\epsilon$ (10 cm) of the target object before the episode ends.
    
    \subsubsection*{\textbf{Experiment 2}} In this experiment, we show that mask-based reward is a viable alternative to the distance-based reward. We trained mask-based GCRL agents with mask-based rewards on the \textit{Franka-MuJoCo} and \textit{UR10e-MuJoCo} environments and compared their performance against the mask-based GCRL agents with distance-based rewards from \textit{Experiment 1 (a)}. The mask-based reward is calculated according to Equation \eqref{eq:reward_mask}, with an added $-1$ reward per step to facilitate faster training. We used the same settings defined in \textit{Experiment 1 (a)} for the two environments. 
    
    \subsubsection*{\textbf{Experiment 3}} 
    In this experiment, we demonstrate that mask-based GCRL agents can learn a complex visual pick-up task from scratch. The objective of the GCRL agent is to grab a target object placed randomly on a table and lift it up 30 cm above the table's surface. We used a modified version of the \textit{UR10e-MuJoCo} environment for this experiment. We used six objects out of 20 for the pick-up targets, which are: apple, green block, donut, toy duck, banana, and white egg. These objects were slightly scaled down in size to facilitate easier grasping. The other 14 objects were used as distractors during training, and in each episode, 2 to 4 objects were randomly sampled and placed on the table.
    
    We compared the performance of mask-based goal conditioning (GC) and mask-based reward system with mask-based GC and distance-based reward system, and 3D position-based GC and distance-based reward system. For the mask-based GC, we selected the part of the image between the gripper fingers as the region of interest (ROI). This is where the object is expected to be during a successful grasp (Figure \ref{fig:pickup}). The agent received the mask-size reward only if the target mask was inside the ROI. This approach allowed the agent to better align the robotic arm's fingers with the target for good grasping, resulting in more successful pick-ups. For all three systems, we included the touch information of the target object and the arm's fingers using a binary vector of size 2. For the mask-based system, once both fingers touched the target object, the mask activation was removed, and only a blank mask was provided. We trained the agents for one million steps, and the first 5000 steps were used to pre-fill the replay buffer, which had a size of 300,000. 
    
    We used the following reward function for this experiment:  
    {
    \begin{equation} \label{eq:reward_pickup}
    R^{\text{pick}} = -1.1 + R^{\text{reach}} + R^{\text{contact}} + R^{\text{lift}} + R^{\text{goal}}
    \end{equation}
    }
    Here, $R^{\text{reach}}$ is set to distance-based reward (Equation \ref{eq:reward_dist}) or mask-based reward (Equation \ref{eq:reward_mask}), and only applied before the agent touches the target with both fingers. $R^{\text{contact}}$ is $10$ for the first time both fingers touch the target and $0.1$ for subsequent touches with both fingers. $R^{\text{lift}}$ is set to $1 - 3.3 \cdot (D)$, where $D$ is the current distance to the target height. $R^{\text{lift}}$ is used after both fingers touch the target object. $R^{\text{goal}}$ is a one-time reward of 10 upon reaching the target height. 
    
    After the training was completed, we used the trained policy to perform 50 pick-up trials for each of the six objects we used during training.
    
    \section{Results}\label{sec:results}
    
    \subsubsection*{\textbf{Experiment 1 (a)}} 
\begin{figure*}[t]
      \centering
      \begin{subfigure}[t]{0.32\linewidth}
        \includegraphics[width=\linewidth]{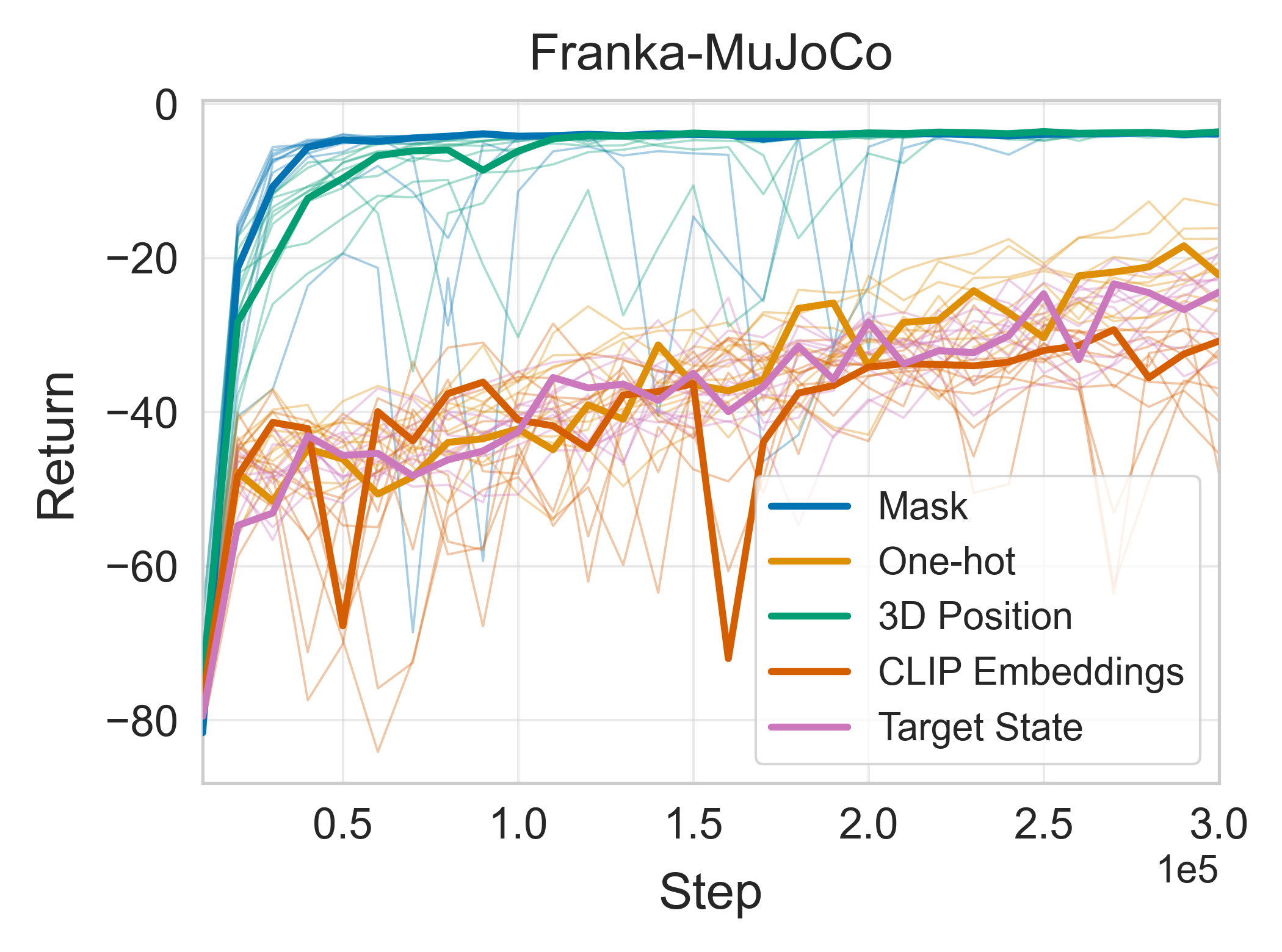}
        \vspace{-1.85em} 
        \caption{\scriptsize Franka-MuJoCo experiments: The mask-based GCRL agent shows faster learning and similar final performance compared to the 3D position-based agent.}
        \label{fig:franka-mujoco}
      \end{subfigure}\hfill
      \begin{subfigure}[t]{0.32\linewidth}
        \includegraphics[width=\linewidth]{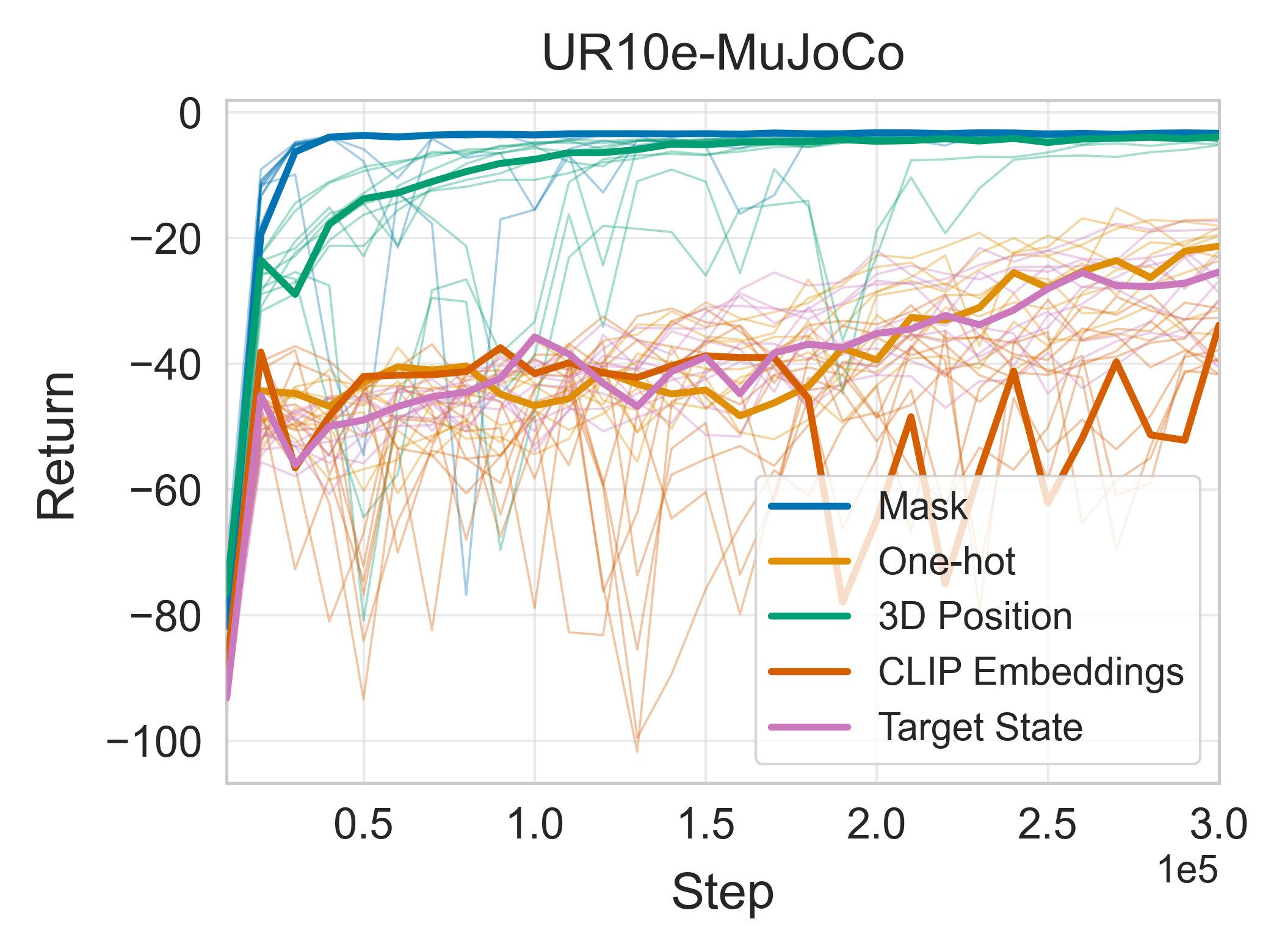}
        \vspace{-2em} 
        \caption{\scriptsize UR10e-MuJoCo experiments: The mask-based GCRL agent shows similar learning and performance behavior as observed in \ref{fig:franka-mujoco}.} 
        \label{fig:ur10e-mujoco}
      \end{subfigure}\hfill
      \begin{subfigure}[t]{0.32\linewidth}
        \includegraphics[width=\linewidth]{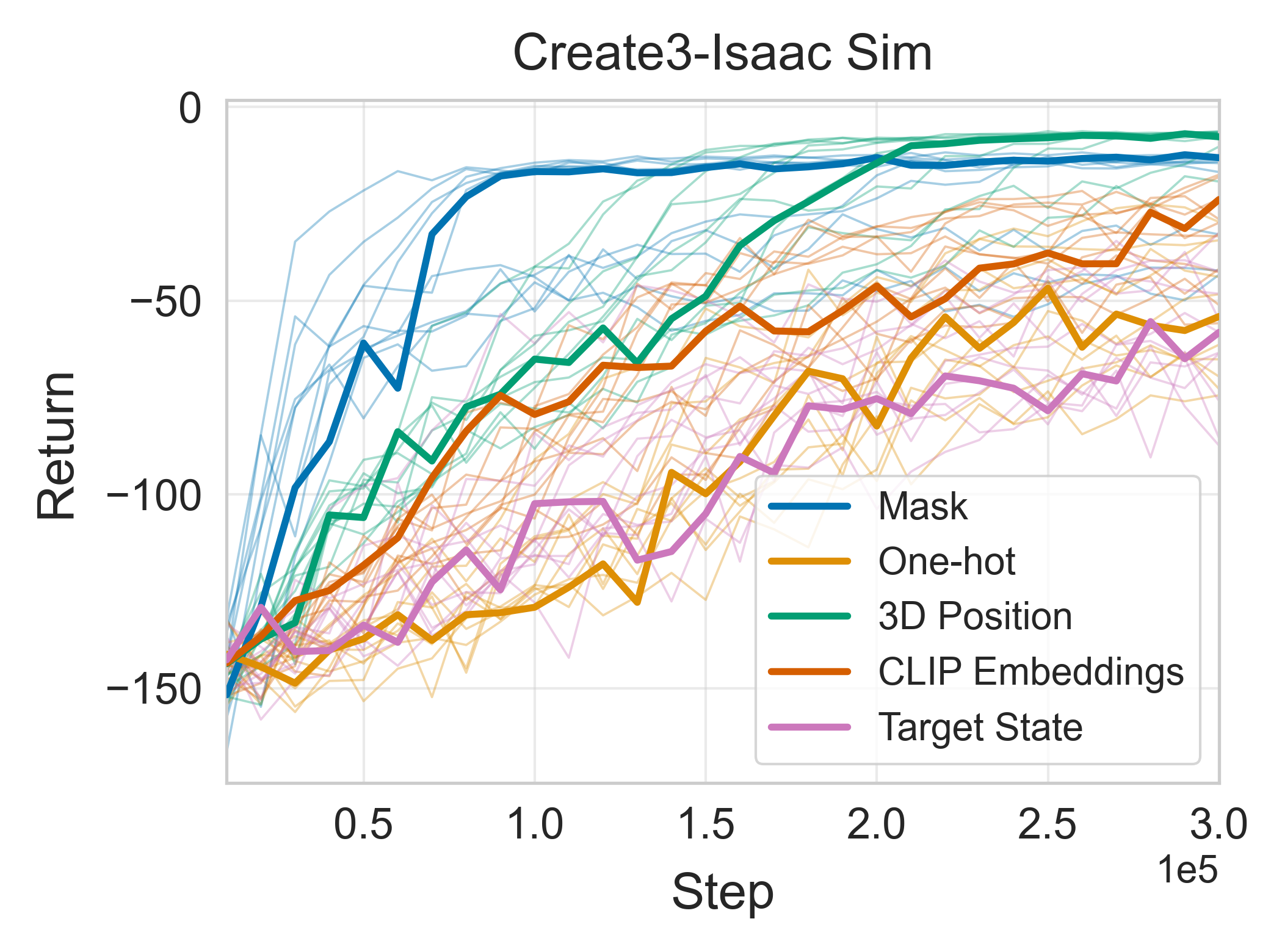}
        \vspace{-2em} 
        \caption{\scriptsize Create3-Isaac-Sim: The median mask-based agent shows faster learning, but the median 3D position-based agent has better final performance due to having access to the privileged target position.}
        \label{fig:Create3-is}
      \end{subfigure}
    
      \vspace{0.4em}
    
      \hspace*{\fill}
      \begin{subfigure}[t]{0.32\linewidth}
        \includegraphics[width=\linewidth]{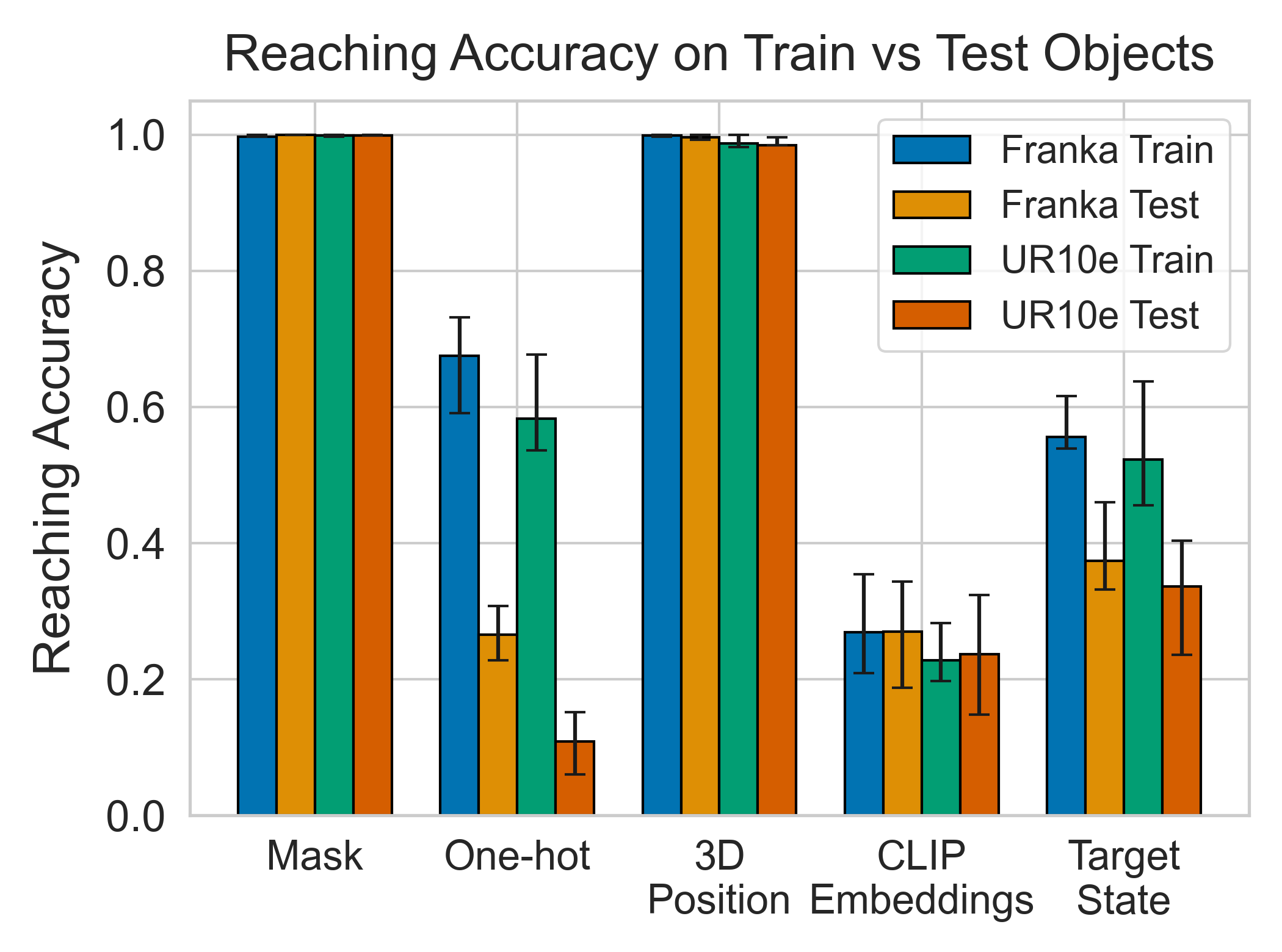}
        \vspace{-1.75em} 
        \caption{\scriptsize Reaching accuracy of trained agents (Franka \& UR10e MuJoCo environments). The mask-based agents have the best reaching accuracy, closely followed by the 3D position-based agents.}
        \label{fig:reach_acc}
      \end{subfigure}
      \hspace{0.06\linewidth}
      \begin{subfigure}[t]{0.32\linewidth}
        \includegraphics[width=\linewidth]{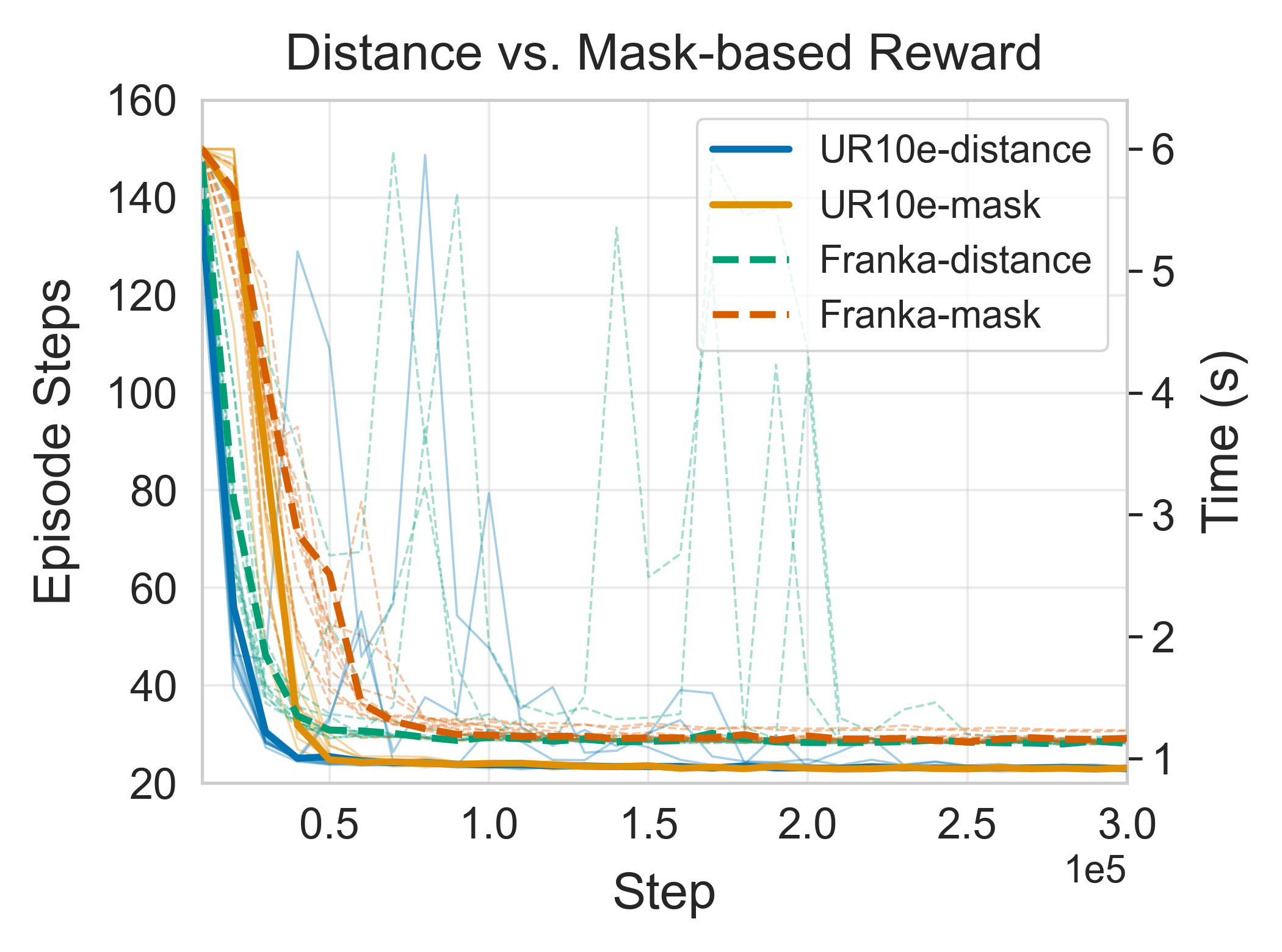}
        \vspace{-1.75em} 
        \caption{\scriptsize Comparison of distance-based vs. mask-based reward systems. The final performance between these systems is similar.}
        \label{fig:reward_comp}
      \end{subfigure}
      \hspace*{\fill}
    
      \caption{Simulated experiment results. Bold lines show the median run (episodic return), and faint lines indicate other seeds.}
      \label{fig:example2}
    \end{figure*}

    Figure \ref{fig:franka-mujoco}, \ref{fig:ur10e-mujoco}, and \ref{fig:Create3-is} show the results for \textit{Experiment 1 (a)}. The bold lines show the learning performance of the median run selected based on the episodic return, and the faint lines represent the remaining seeds following \citeauthor{tanaka2026performance} (\citeyear{tanaka2026performance}). In all three experiments, the mask-based system achieved the fastest learning performance. For the \textit{Franka-MuJoCo} and the \textit{UR10e-MuJoCo} runs, the mask-based GCRL agent learned the optimal policy similar to the 3D position system. For the \textit{Create3-Isaac Sim} experiments, the overall performance of the mask-based approach was slightly lower than the 3D-position method. This is because, for the 3D position, the agent always had access to the privileged target position, allowing it to learn to reach the target directly, even when not in view. On the other hand, for the mask-based method, the agent learns two skills: first, to find the target by rotating, and second, to move close to the target. Approaches such as one-hot encoding, CLIP embeddings, and target state showed poor learning performance, revealing the importance of selecting the appropriate goal conditioning method to achieve fast policy convergence.

    \subsubsection*{\textbf{Experiment 1 (b)}}
    Figure \ref{fig:reach_acc} shows the results for \textit{Experiment 1 (b)}. The bars in the plot represent the median reaching accuracy, and the whiskers represent the first and third quartiles. The mask-based GCRL agents achieved the highest performance for train and test objects, with 99.9\% average accuracy for both. 3D position, on the other hand, achieved 98.9\% and 98.8\% average accuracy for train and test, respectively. This result shows that the agents learned to utilize the masks effectively for both train and test objects. 
    
    \subsubsection*{\textbf{Experiment 2}} In Figure \ref{fig:reward_comp}, we used average episode steps instead of return for comparison, as the distance-based and mask-based rewards are not directly comparable. The median runs are selected based on average episodic steps. The mask-based reward system performed similarly to the distance-based reward system while having higher stability during training. This result shows that the mask-based reward system is a viable alternative to the distance-based reward. 
    
    \subsubsection*{\textbf{Experiment 3}} 
\begin{figure*}[t]
      \centering
      \begin{subfigure}[t]{0.32\linewidth}
        \includegraphics[width=\linewidth]{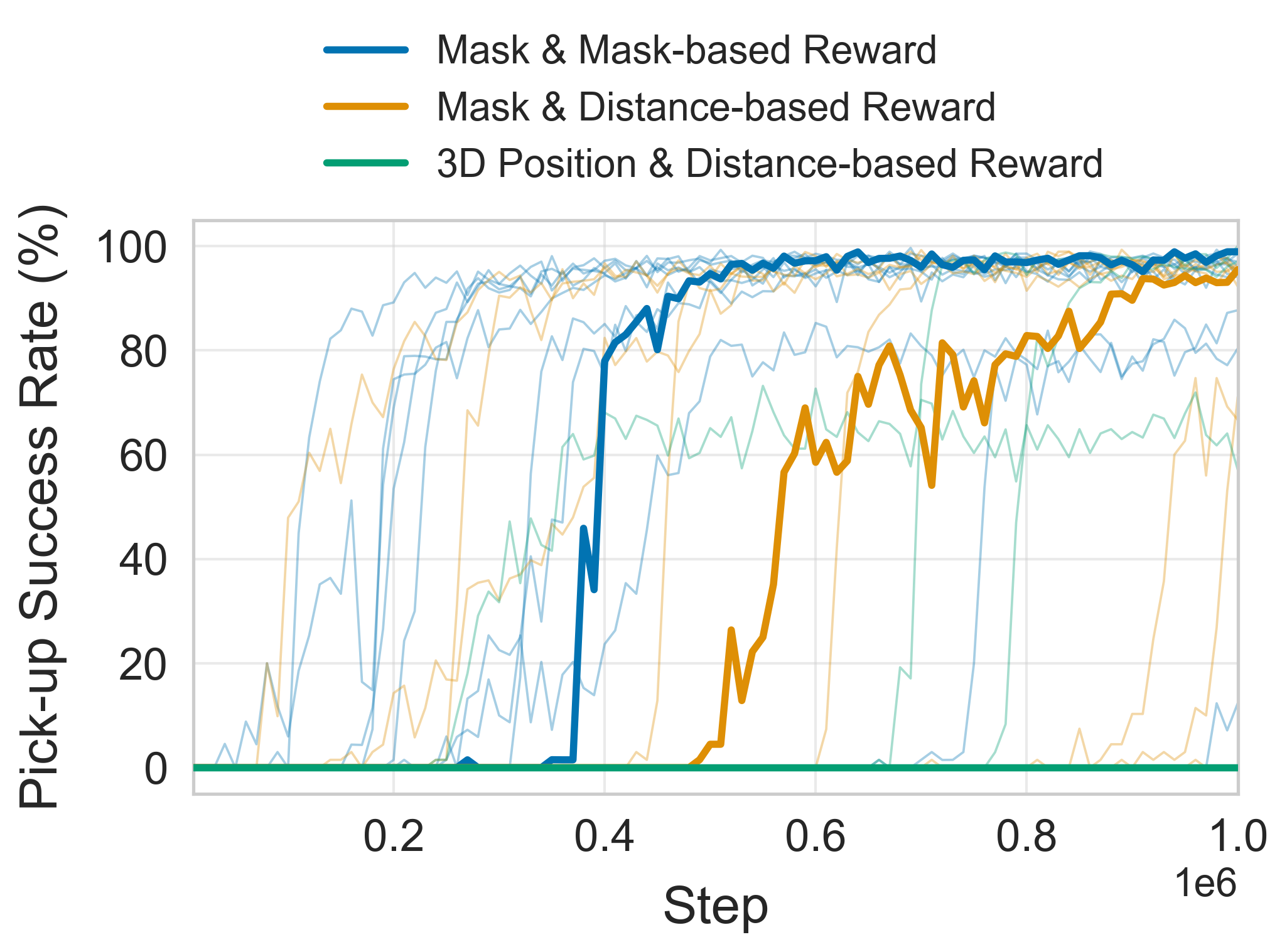}
        \vspace{-1.75em} 
        \caption{\scriptsize Pick-up success rate during training using the UR10e arm. The mask-based systems performed better than the 3D-position-based system.} 
        \label{fig:pickup_lc}
      \end{subfigure}\hfill
      \begin{subfigure}[t]{0.32\linewidth}
        \includegraphics[width=\linewidth]{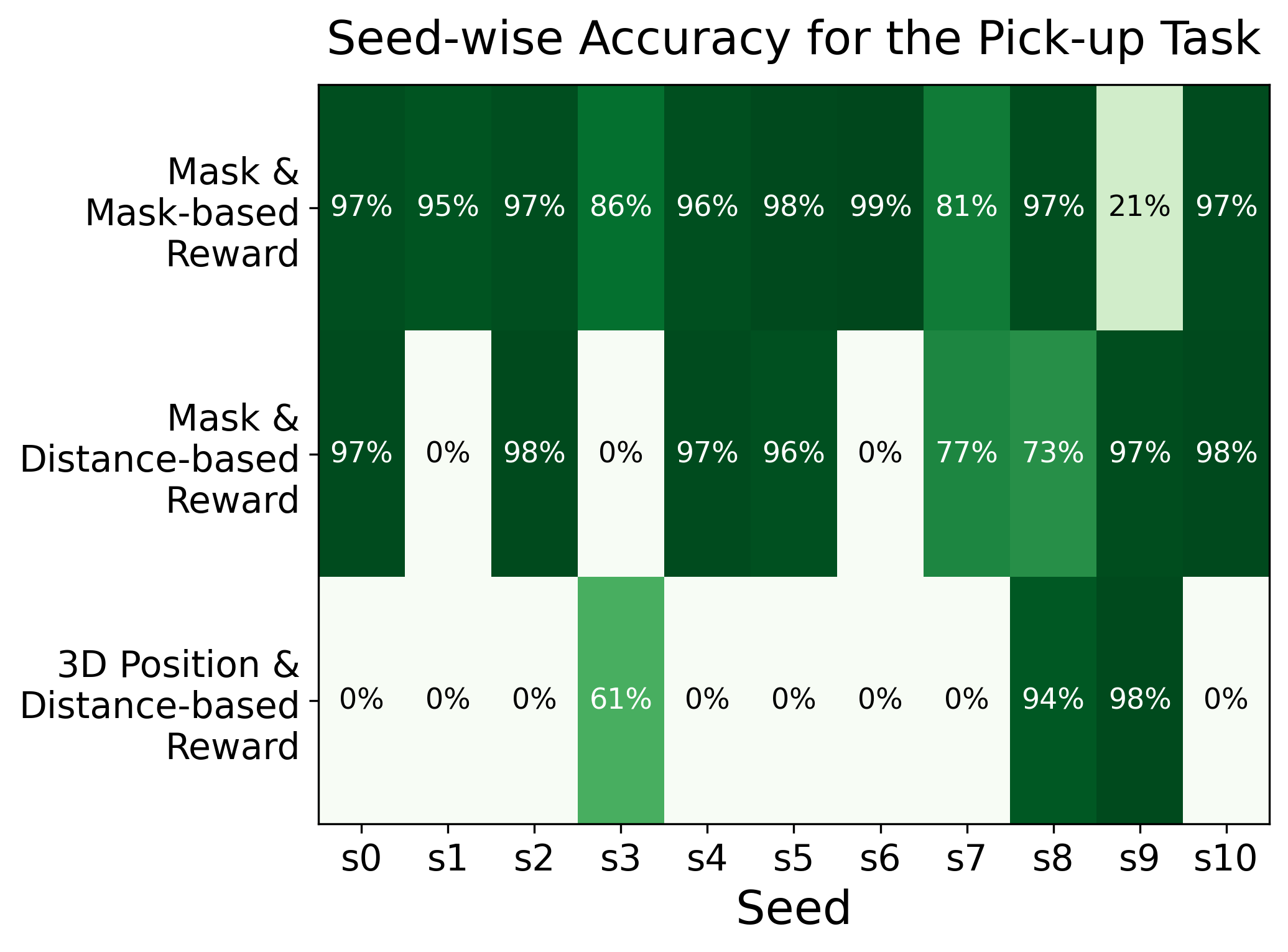}
        \vspace{-1.85em} 
        \caption{\scriptsize Success rate of the trained agents in the post-training trials.}
        \label{fig:seed_acc1}
      \end{subfigure}\hfill
      \begin{subfigure}[t]{0.32\linewidth}
        \includegraphics[width=\linewidth]{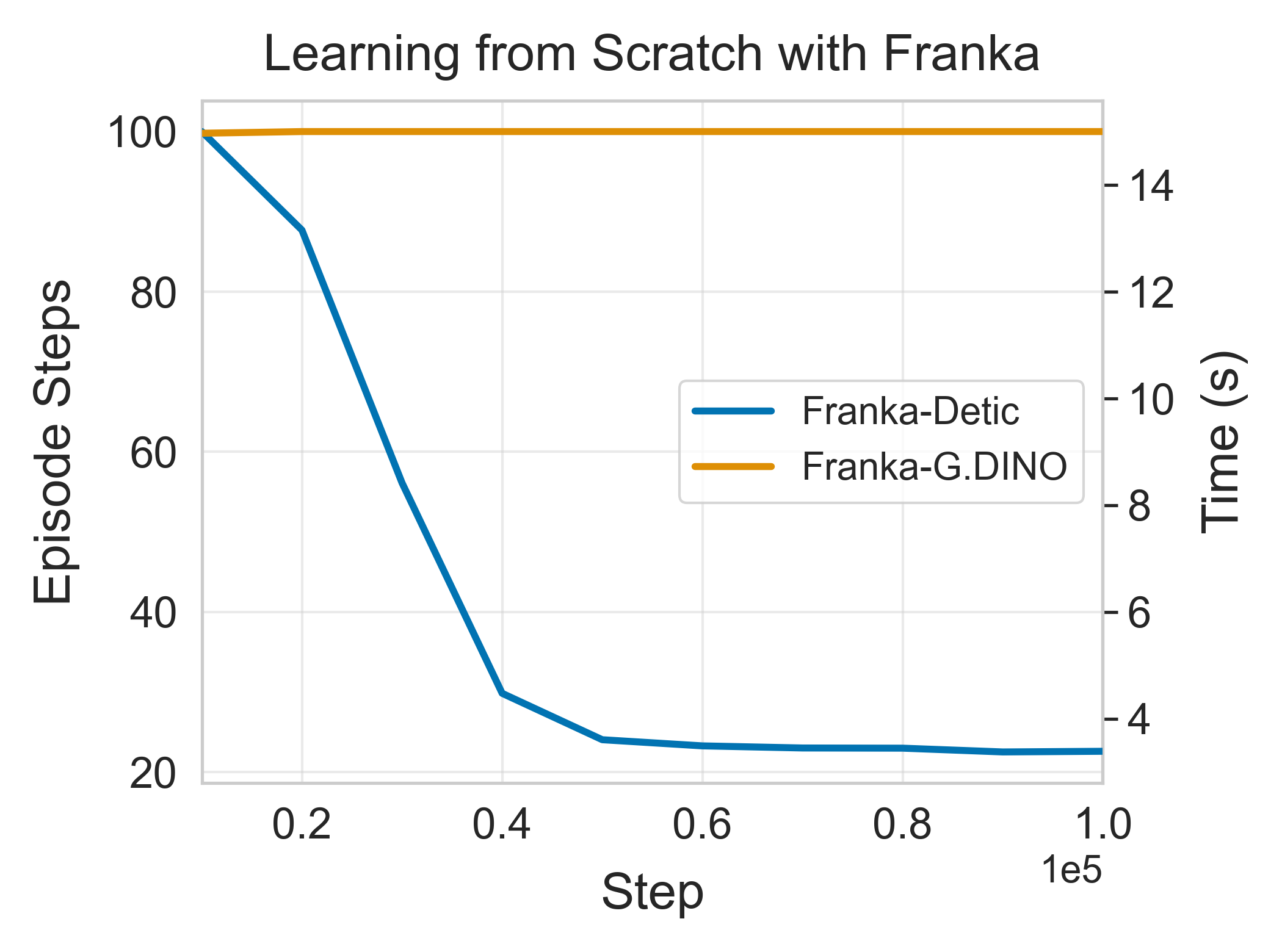}
        \vspace{-1.75em} 
        \caption{\scriptsize Real-world experiments with Detic and Grounding DINO. For our real-world setup (\ref{fig:franka_rw}), Detic-based agent performed much better than Grounding DINO.}
        \label{fig:real_world_exp_from_scratch}
      \end{subfigure}
    
      \caption{ (a) Pick-up success rate during training (UR10e MuJoCo), where bold lines show the median runs. (b) Post-training success per seed (50 trials per object). (c) Learning curves for the real-world learning from scratch task.}
    \end{figure*}

\begin{figure*}[t]
        \centering
        \includegraphics[
        width=0.9\linewidth]{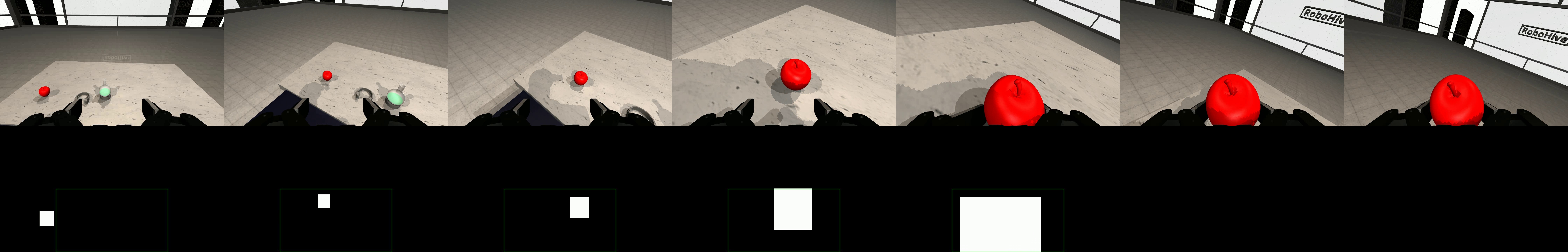}
    
        \setlength{\belowcaptionskip}{-5pt}
        
        \caption{ A trained mask-based GC and reward agent successfully picks up the apple object. The green rectangles in the masks (bottom images) represent the ROI (shown only for illustration). In the last two frames, the agent lifts up the apple object after a successful grasp.} 
        \label{fig:pickup}
        
    \end{figure*}

    This experiment examines whether mask-based goal conditioning supports learning a complex pick-up task across diverse object shapes and sizes. Figure \ref{fig:pickup_lc} shows the pick-up accuracy during training, and Figure \ref{fig:seed_acc1} shows the pick-up accuracy of the trained agents. Nine out of eleven seeds for the mask-based GC and rewards system achieved over 90\% pick-up success. On the other hand, only two seeds for the 3D position-based GC and distance-based reward system achieved over 90\% success. This result shows that the ROI based masks guided the arm to a good grasp position (Figure \ref{fig:pickup}), which resulted in higher pick-up accuracy. The dense distance-based reward (Equation \ref{eq:reward_dist}) provides no incentive for the agent to align its end-effector with the target for grasping. In the failed runs, the agent hovered near the target object but did not learn to grasp or lift it. This experiment demonstrates that the mask-based system can be trained to perform complex visual pick-up tasks from scratch and shows the robustness of our proposed system. 
    
    \section{Real-world Applications} \label{sec:real_world_app}
    We demonstrate the real-world applicability of our system through two applications: sim-to-real transfer and learning from scratch on physical robots.

    \subsection{Sim-to-Real Application with UR10e} \label{sec:Sim2Real}
\begin{figure*}[ht]
      \centering
      \begin{subfigure}[t]{0.205\linewidth}
        \includegraphics[width=\linewidth]{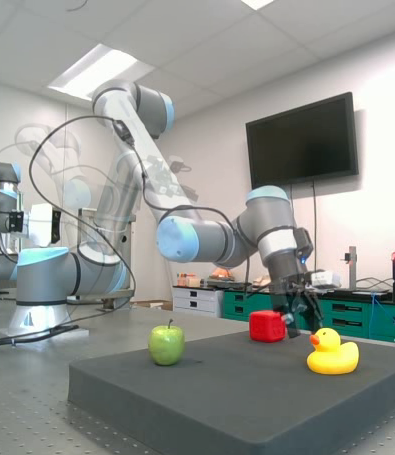}
        \caption{\scriptsize UR10e arm reaching for the rubber duck, controlled using a zero-shot sim-to-real GCRL agent.}
        \label{fig:ur10e_reaching}
      \end{subfigure}\hfill
        \begin{subfigure}[t]{0.28\linewidth}
        \centering
        \includegraphics[width=\linewidth]{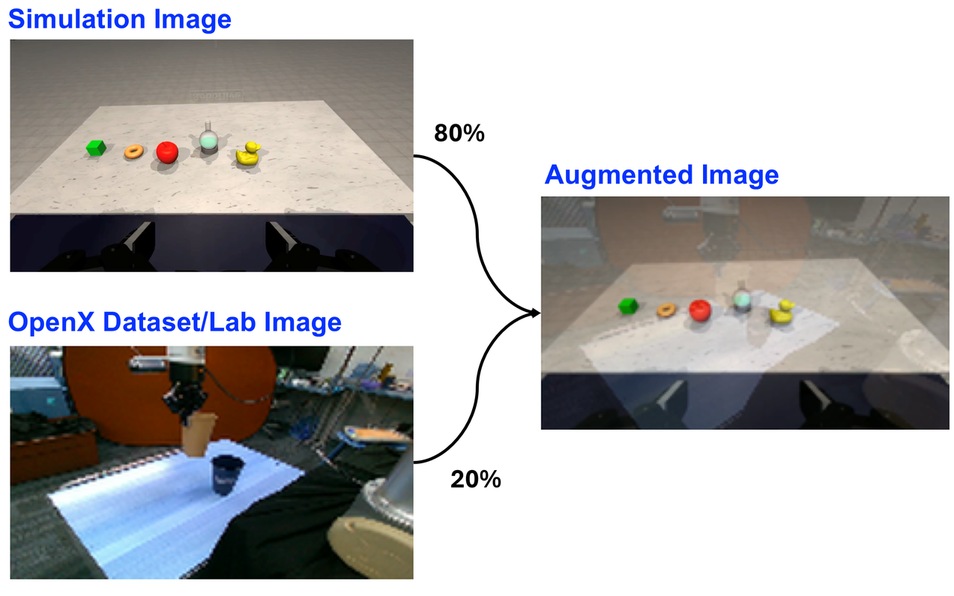}
        \caption{\scriptsize Image augmentation during policy training mixed 80\% of a simulation image with 20\% of a random real-world image.}
        \label{fig:image_mixing}
      \end{subfigure}\hfill
      \begin{subfigure}[t]{0.188\linewidth}
        \includegraphics[width=\linewidth]{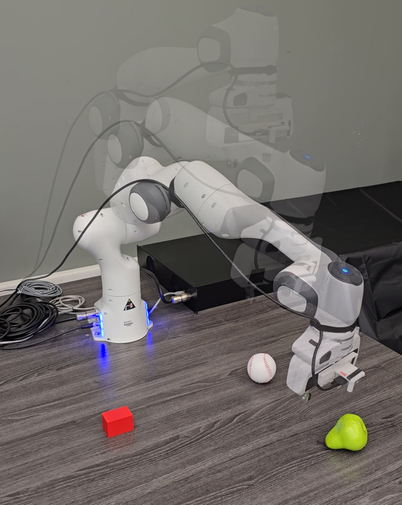} 
        \caption{\scriptsize Franka arm reaching for a pear in the learning from scratch task using Detic.}
        \label{fig:franka_reaching}
      \end{subfigure}\hfill
      \begin{subfigure}[t]{0.21\linewidth}
        \includegraphics[width=\linewidth]{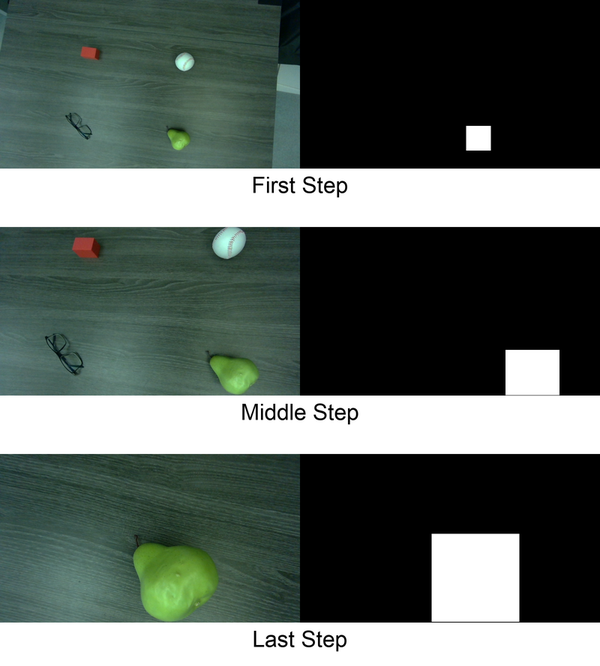}
        \caption{\scriptsize Franka camera images and masks created using Detic at different stages of the episode to reach the pear.}
        \label{fig:fr_cam_stacked}
      \end{subfigure}\hfill
      \setlength{\belowcaptionskip}{-6pt}
      \caption{Sim-to-real using UR10e and real-world learning from scratch using Franka.}
      \label{fig:fig_four}
    \end{figure*}

    To assess the real-world applicability of our mask-based GCRL system, we performed a sim-to-real transfer of the proposed algorithm onto a physical UR10e robot (Figure \ref{fig:ur10e_reaching}). The policy was trained using PPO in the \textit{UR10e-MuJoCo} environment with the addition of domain randomization. We trained PPO with a custom multi-input policy using three-frame stacking and four parallel environments. We used ground-truth masks in simulation, where the privileged target information is readily available, and utilized Grounding DINO for mask generation in the real world. Grounding DINO uses ($480 \times 848$) resolution images from a wrist-mount RealSense D435 camera. These are rescaled to $120 \times 212$ as input to the PPO CNN network. The best-performing policy from 10 seeds was selected for zero-shot transfer. The agent operates at a wall-clock time step of about 120 ms, dominated by Grounding DINO inference. A background control thread sends \texttt{servoJ} commands at 500 Hz, and a successful reach takes under 30 seconds.

    Domain randomization techniques are used to mitigate the discrepancies between simulated environments and the real world. The UR10e robot's joint readings received Gaussian noise $\sim \mathcal{N}(0, 0.05)$ while its initial qpos varied by $\pm 0.2$ for the first 5 DoFs. RGB images were augmented with contrast, saturation, brightness, and Gaussian blur ($\sigma$), all randomized within [0.8, 1.2]  using Kornia \citep{eriba2019kornia}. As shown in Figure \ref{fig:image_mixing}, we introduce a novel image augmentation process to reduce the image domain shift by linearly interpolating simulation images with randomly selected images from diverse real-world robotics tasks during training. This image augmentation was found to be essential for policy transfer.

    We evaluate our system using in-distribution objects (green apple, red cube, yellow rubber duck) and two out-of-distribution test sets (Demo 1: toy robot, frog, water bottle; Demo 2: toy robot, plush dinosaur, water bottle). These objects were selected to provide variation in mass, size, color, shape, and surface texture, testing whether the reaching behavior learned in simulation generalizes to visually diverse real-world objects. The objects were placed at pre-selected positions whose coordinates in the robot's base frame were measured in advance. The distance in each trial is computed automatically from the final gripper midpoint to the recorded object position.
    
    \begin{table}[htbp]
    \centering
    \caption{\small Performance Comparison Between In-Distribution and Out-of-Distribution (OOD) Objects. Results obtained with a frame skip of 4 on the UR10e, 25 trials per object set. We classify a trial as successful if the end effector reaches within 15 cm of the target object.}
    \label{tab:results}
    \begin{tabular}{@{}lccc@{}}
    \toprule
    \textbf{Metric} & \textbf{In-Distribution} & \textbf{OOD (Demo1)} & \textbf{OOD (Demo2)} \\
    \midrule
    Success Rate (\%) & 96 & 92 & 92 \\
    Distance (m) & $0.076 \pm 0.035$ & $0.13 \pm 0.07$ & $0.092 \pm 0.050$ \\
    \bottomrule
    \end{tabular}
    \end{table}
    
    Table~\ref{tab:results} demonstrates the successful real-world transfer of the policy, achieving success rates comparable to simulation with ground-truth masks (89\% in-distribution, 90\% OOD), indicating no degradation from sim-to-real transfer. The end-effector can reach within 15 cm of the target objects, regardless of their size and location.

    We performed two additional tasks in the real world using the pretrained policy. We conducted pick-and-place by coupling a heuristic grab, lift, and place mechanism with the reaching policy. We also performed a tracking task by placing a target object in front of the arm and moving it gradually. The agent executed both tasks successfully, showing the robustness of the pretrained policy and Grounding DINO's ability to generate masks for GCRL in the real world.

    \subsection{Learning from Scratch with Franka}
    
    We show the application of the mask-based system (goal representation and reward) in a real-world learning from scratch task using the Franka Panda arm (Figure \ref{fig:franka_rw}). We simplified the reaching task in the real world by using four stationary objects placed on a table in front of the arm: a red block, a baseball, a pear (artificial), and eyeglasses. We used Detic and Grounding DINO models for object detection and used slightly lower resolution images ($450\times 800$) to facilitate faster inference times. We used 100 steps per episode and trained for 100,000 steps. The termination distance, $\epsilon$, was set to 13 cm from the target.  We used an Intel RealSense D405 camera attached near the end-effector of the arm to capture images.  We performed the training on a computer equipped with a Ryzen Threadripper 3970X processor and an Nvidia RTX 3090 GPU. We used the same SAC GCRL system that we used in the simulated experiments. The GCRL agent had to learn to reach the randomly selected target object by controlling the joint velocities of the arm. Average step times were 157 ms with Detic (131 ms inference) and 168 ms with G.DINO (158 ms inference). Total run durations were 449 and 338 minutes, respectively. The Detic agent completed more episodes (3,328 vs. 1,000) and therefore spent more time on arm resets, which take about 3 seconds each.
    
    Figure \ref{fig:real_world_exp_from_scratch} shows the results of the real-world learning from scratch experiments. The Franka-Detic agent learned an optimal reaching policy in 60,000 steps. Figure \ref{fig:franka_reaching} shows the trained agent reaching for a pear, and Figure \ref{fig:fr_cam_stacked} shows the associated camera images and masks generated using Detic. On the other hand, the Grounding DINO model had severe false-positive detection issues. The agent failed to reach the target objects as it tried to exploit rewards from false detection masks. This experiment shows the possibilities and difficulties of using pretrained object detection models in real-world learning from scratch tasks. 
    
    We further evaluated the trained Detic-based GCRL agent by performing a tracking task with an unseen object. We placed a plastic banana on top of the table and moved it using a stick while the GCRL agent was controlling the arm. The agent managed to follow the banana easily, even though it was trained on reaching stationary objects. The successful tracking behavior demonstrates the robustness and the generalization capabilities of our proposed system.

    \section{Limitations and Future Work}
    Our results demonstrate that dynamic object masks provide a simple and effective goal representation and dense reward signal for visual GCRL. We outline two directions for future work that would extend the scope of this paper.

     \textbf{Mask quality and robustness.} In this work, we used bounding-box-based masks as the goal representation. We experimented with segmentation masks as a finer alternative, but observed a slight performance degradation, likely due to reduced generalization across objects, but this requires further investigation. We note that the mask by itself enables an object-agnostic \emph{reaching} skill rather than a fully goal-agnostic manipulation skill. In addition, the bounding box is agnostic to the nuances of the object. Downstream tasks (after reaching) involving grasping or picking may require reaching for, say, the handle of the cup instead of the middle of the cup. However, the agent also has access to information apart from the mask, such as the RGB camera images, which can help the agent learn to perform downstream tasks.
     The performance of our system is upper-bounded by the quality of the object detector, as observed in our real-world experiments (Figure \ref{fig:real_world_exp_from_scratch}). However, the detector is a replaceable component, and our approach benefits directly from improved detection models. A useful next step is to study how sensitive the proposed system is to mask errors, such as missing regions, flickering detections, partial occlusions, noise, and false detections, and measure the impact on performance. We note that improving mask quality is more closely related to image processing than GCRL, and we leave a thorough investigation to future work.
    
    \textbf{Sim-to-real and learning from scratch in more challenging environments.} We primarily evaluated indoor tabletop manipulation and navigation scenarios, where our image interpolation technique using indoor real-world data helped bridge the simulation-to-real domain gap, and learning from scratch was feasible due to the controlled nature of the setup. An important next step is to extend mask-conditioned GCRL to more challenging settings, including outdoor environments and multi-task settings such as navigation followed by manipulation, sequential pick-and-place, or outdoor mobile manipulation tasks (e.g., weeding or fruit harvesting, such as strawberry picking). For sim-to-real, this direction requires building high-quality outdoor simulation setups and collecting diverse outdoor real-world data for image interpolation, which is substantially more challenging than our current indoor setup. For learning from scratch, unstructured outdoor environments introduce additional challenges for the GCRL agent, such as a more complex state space, movement limitations, and safety concerns, which we aim to address in future work.

    \section{Conclusion}
    
    We introduced an object-agnostic mask-based GCRL and reward system. Experimental results suggest that our proposed system     performs similarly to or better than other approaches on visual manipulation and reaching tasks. The mask-based systems showed superior reaching performance on novel objects utilizing RGB images only, achieving 99\% accuracy.  Moreover, we showed that dense reward based on mask size is an effective alternative to distance-to-target-based reward, diminishing the necessity for distance calculation methods. The mask-based agents learned successful pick-up behavior, with nine of eleven seeds exceeding 90\% success. Finally, we showed that our proposed method is effective in real robot learning, and open-vocabulary object detection models are effective when transferring trained policies from simulation to real robots.

    \section*{Acknowledgments}
    We thank the reviewers for their valuable feedback. We gratefully acknowledge the National Research Council of Canada's (NRC) AI for Design Challenge Program and the Canada CIFAR Chairs program for their financial support. We also thank the Digital Research Alliance of Canada for providing computational resources.
    
    \bibliography{ref}
    \bibliographystyle{rlj}
    
    \end{document}